\documentclass[preprint,12pt, a4paper]{elsarticle}

\usepackage{amssymb}
\usepackage{hyperref}
\usepackage{subcaption}

\journal{SoftwareX}

\begin{document}
\renewcommand{\labelenumii}{\arabic{enumi}.\arabic{enumii}}

\begin{frontmatter}

\title{MitoSeg: Mitochondria Segmentation Tool}


\author[label1]{Faris Serdar TAŞEL}
\author[label2]{Efe ÇİFTCİ}
\address[label1]{Department of Computer Engineering, Çankaya University, Ankara, Turkey, fst@cankaya.edu.tr}
\address[label2]{Department of Computer Engineering, Çankaya University, Ankara, Turkey, efeciftci@cankaya.edu.tr}

\begin{abstract}
Recent studies suggest a potential link between the physical structure of mitochondria and neurodegenerative diseases. With advances in Electron Microscopy techniques, it has become possible to visualize the boundary and internal membrane structures of mitochondria in detail. It is crucial to automatically segment mitochondria from these images to investigate the relationship between mitochondria and diseases. In this paper, we present a software solution for mitochondrial segmentation, highlighting mitochondria boundaries in electron microscopy tomography images and generating corresponding 3D meshes.
\end{abstract}


\end{frontmatter}


\section*{Metadata}

\begin{table}[!ht]
\begin{tabular}{|l|p{6.5cm}|p{6.5cm}|}
\hline
\textbf{Nr.} & \textbf{Code metadata description} & \textbf{Please fill in this column} \\
\hline
C1 & Current code version & v1.0 \\
\hline
C2 & Permanent link to code/repository used for this code version & \url{https://github.com/fstasel/mitoseg} \\
\hline
C3  & Permanent link to Reproducible Capsule & \url{https://github.com/fstasel/mitoseg/tree/master/docker}\\
\hline
C4 & Legal Code License   & GPLv3 \\
\hline
C5 & Code versioning system used & git \\
\hline
C6 & Software code languages, tools, and services used & C++, CMake, Docker \\
\hline
C7 & Compilation requirements, operating environments \& dependencies & A modern GNU/Linux system with development files for OpenCV, Boost, and yaml-cpp libraries installed \\
\hline
C8 & If available Link to developer documentation/manual & \url{https://github.com/fstasel/mitoseg#readme} \\
\hline
C9 & Support email for questions & fst@cankaya.edu.tr \\
\hline
\end{tabular}
\caption{Code metadata (mandatory)}
\label{codeMetadata} 
\end{table}



\section{Motivation and significance}
Mitochondria are organelles responsible for producing the chemical energy required for various biochemical reactions in the cell. The relationship between mitochondria and neurodegenerative diseases such as Alzheimer's and Parkinson's has become an area of increasing interest, as understanding the causes of these diseases is of great importance \cite{payne2024multimodal,pradeepkiran2024mitochondria,reiss2024mitochondria,borsche2021mitochondria,malpartida2021mitochondrial,monzio2020role,wang2020mitochondria}. For this reason, it is essential to study the physical structure of mitochondria.\\

Advances in electron microscopy imaging techniques have significantly impacted the investigation of subcellular structures. Among these techniques, Serial Block-Face Scanning Electron Microscopy (SBFSEM) and Transmission Electron Microscopy (TEM) are frequently used, offering detailed imaging down to a few nanometers \cite{tasel2016validated}. Such high-resolution imaging allows for the observation of mitochondrial membrane structures, including the boundary and internal regions. Since mitochondria can exist in a condensed form (where the internal structure is packed with proteins), they often appear as dark blob-like objects in electron microscopy images. Due to this characteristic, the specimen must undergo a pre-processing step known as heavy metal staining, which enhances the visibility of the internal arrangement known as cristae, allowing for detailed examination and investigation \cite{perkins2003three}. Following the necessary preparation steps, the crista structure must be scanned at a resolution high enough to reveal its fine details.\\

In existing literature, various mitochondria segmentation methods that utilize different modalities are available \cite{nevsic2024automated,franco2022stable,li2022advanced,somani2022virtual,fischer2020mitosegnet,mekuvc2020automatic,peng2020unsupervised,xiao2018automatic,lucchi2011supervoxel}. These methods generally utilize deep learning approaches and CNNs and require ground truth data for training purposes. As these studies achieve significant success in mitochondria segmentation, they frequently have focused on condensed mitochondria. However, it should be noted that these segmentation algorithms should work with mitochondria images that are also suitable for crista analysis.\\

MitoSeg is a tool developed for mitochondria detection and segmentation based on the algorithm proposed in \cite{tasel2016validated}, which is designed to work on datasets exhibiting clear cristae structures. This method enables the segmentation of mitochondria from the Electron Microscopy Tomography (EMT) images using pre-processed specimens mentioned above by leveraging the general physical characteristics of mitochondria without the need for a training phase. Furthermore, the method is adaptable to mitochondrial images obtained from different imaging modalities. For MitoSeg to produce results, a high-resolution intracellular image dataset composed of a set of slices and the corresponding metadata (e.g., slice range and pixel size) is required.

\section{Software description}
MitoSeg is a command line utility that works with EMT images. It reads through a set of EMT images and produces 2D image and 3D mesh outputs in which the detected mitochondria boundaries are highlighted. It is developed in C++ and uses the following libraries to operate:

\begin{itemize}
 \item \textbf{OpenCV4:} OpenCV handles the fundamental image processing tasks.
 \item \textbf{Boost:} The Boost library handles string manipulation and command line options.
 \item \textbf{yaml-cpp:} MitoSeg is developed with pre-tuned internal segmentation settings, but it is also designed to allow the users to override these settings via external sources without recompiling MitoSeg. The YAML file format is chosen for its simple syntax among many existing standard formats. The yaml-cpp library provides easy-to-use programming capabilities for loading the user-defined settings from these external files.
\end{itemize}

\subsection{Software architecture}
\begin{figure}[ht]
\centering
\includegraphics[width = \linewidth]{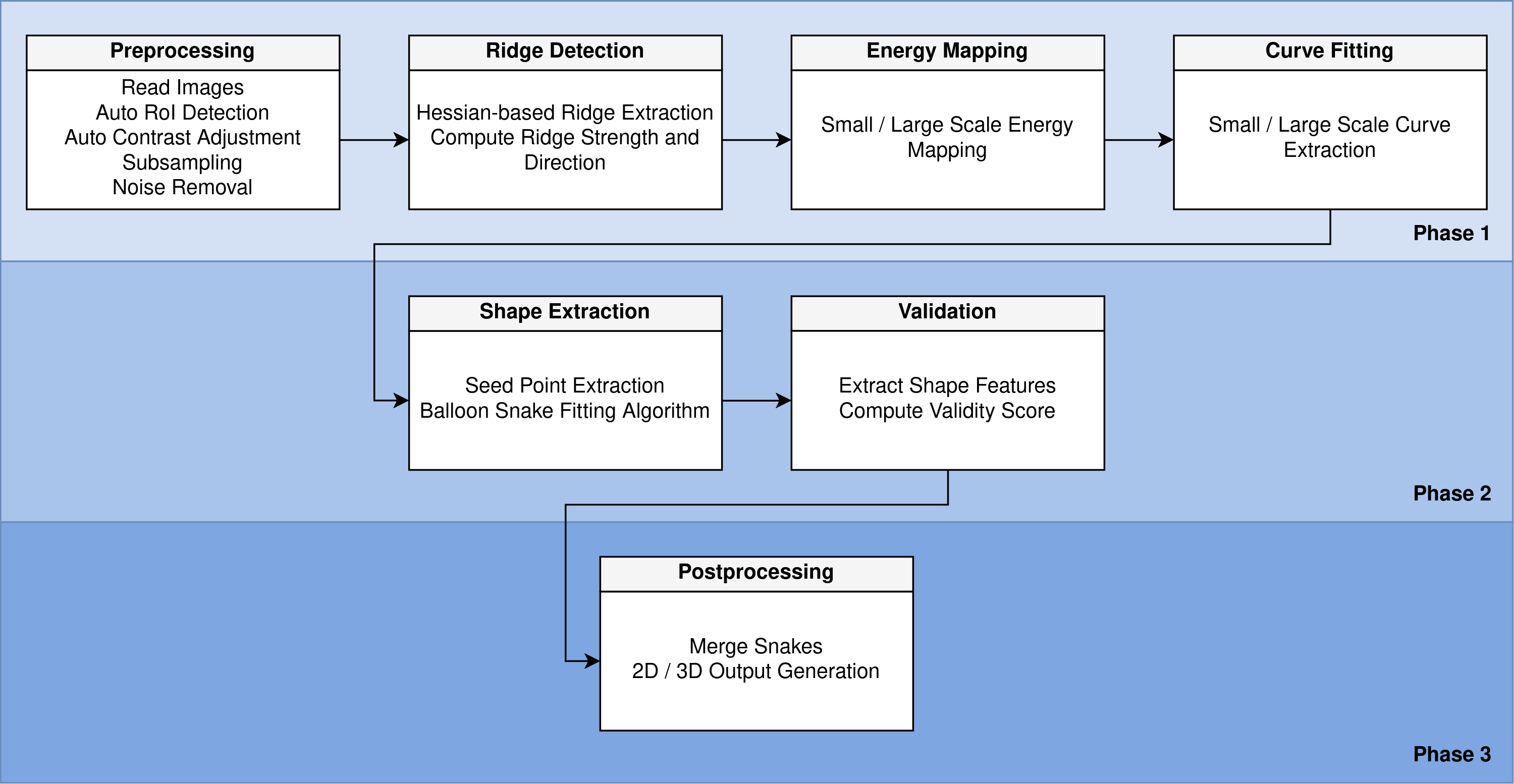}
\caption{Flowchart of the algorithm.}\label{fig:fig1}
\end{figure}

The software runs in three separate phases, each containing multiple substeps, as illustrated in Figure \ref{fig:fig1}. The following sections summarize each phase.

\subsubsection{Phase 1 - Preprocessing, Ridge Detection, Energy Mapping, Curve Fitting}
This phase handles the preprocessing of provided dataset images and generates intermediate data required by the actual segmentation process. Since EMT of mitochondria can be a set of reconstructed images obtained from a preprocessed specimen, it may contain unsharp borders, have a low-contrast intensity distribution, and some artifacts involving extreme high and low-intensity levels. The first step of the preprocessing defines a region of interest automatically (if the user does not provide it) by cropping the image from the borders that do not contain useful information. In the second step, an auto-contrast adjustment method is employed in order to remove the extremity in the image histogram and normalize intensity values into 0 - 255. Then, the input images are subsampled to 2nm per pixel in the third step to facilitate the parameter tuning. In the last step of the preprocessing, bilateral and Gaussian filtering are applied to input images to emphasize membrane structures while eliminating unwanted noise.\\

The preprocessed images are then used in a Hessian matrix-based ridge detection process to locate membrane-like structures. Note that membranes can be elongated, such as in the periphery of the mitochondrion or relatively short curvy structures (e.g., cristae). In order to distinguish between the two, an energy mapping process is utilized on large and small scales individually, which provides valuable information on membranes such as curvature, strength, and orientation. Then, a curve fitting process that uses a parabolic arc model is employed to extract small and large-scale curve segments. Experiments show that extracted curve segments in different scales are useful for locating the boundary and internal structures of mitochondria.

\subsubsection{Phase 2 - Shape Extraction, Validation}
In this phase, extracted curve segments are used to segment mitochondrial regions. First, seed points are located near curve segments by implementing a modified Density-Based Spatial Clustering of Applications with Noise (DBSCAN) algorithm. Then, seed points are fed to a pseudo-3D balloon snake, which is a variant of an active contour model. This model extracts potential closed regions bounded by the mitochondrial membranes.\\

However, it is also possible to erroneously segment a region in a different organelle (e.g., somewhere in the endoplasmic reticulum) due to the initialization of the model outside of mitochondria. To overcome this problem, a validation step is performed to filter out incorrect segmentations. A validator function checks several properties of the candidate segmentation, such as the average energy of the boundary and internal region, area of the segmented region, discontinuity, curvature, and signature of the contour. Note that those criteria have been formed by the general physical properties of mitochondria appearing in tomograms. All segmentations identified as valid by the validator function are conveyed to the next phase.

\subsubsection{Phase 3 - Postprocessing}
In this phase, valid segmented regions are merged if they overlap. Finally, boundary points on segmented and merged regions are utilized to generate a 3D mesh as a final product packed in IMOD and PLY file formats so that they can be easily visualized and processed for further applications.\\

The algorithm outlined here is a simplified explanation. A more detailed description of the algorithm, along with further discussion and various examples, can be found in \cite{tasel2016validated}.

\subsection{Software functionalities}
MitoSeg provides several functionalities that enhance the proposed tool's overall performance and usability, such as command line arguments, additional algorithm settings, multi-threaded execution, and a Docker environment for running MitoSeg on all major OS families. The details of each functionality are described below.

\subsubsection{Mandatory and Optional Command Line Arguments}
MitoSeg offers the users various command line options to alter the segmentation process. These options are presented to the users in two forms: mandatory and optional arguments.\\

Below is the list of mandatory command line arguments:
\begin{itemize}
 \item \textbf{pattern:} Filename pattern for each slice that can be iterated with C-style formatting (e.g., ``\verb|mito%03d.tif|" can expand into a range of files from mito000.tif to mito999.tif).
 \item \textbf{psize:} Pixel size in nm/px.
 \item \textbf{zrange:} Range of slice numbers to be processed (e.g., \verb|--zrange 40 80|)
\end{itemize}
Following is the list of available command line options:
\begin{itemize}
 \item \textbf{src:} Path of source images. Images belonging to the same dataset in this directory must follow the same naming pattern (e.g., slice0001.bmp to slice0200.bmp or mito1.tif to mito100.tif). Source images will be looked for in the current working directory if not specified.
 \item \textbf{dst:} Path of the directory where the intermediate and final output files are stored. If not specified, outputs will be stored in the current working directory.
 \item \textbf{roi:} Region of interest in the provided slices (left, top, width, height). If not specified, RoI will be calculated automatically. Manually specifying the RoI can improve performance and help produce better results by focusing on a specific area in a high-resolution image set.
 \item \textbf{phase:} By default, MitoSeg runs the previously explained phases in succession. This option allows for individual running of only a specified phase.
 \item \textbf{valid:} Specifies the threshold of the validator function between 0 and 1. If not specified, it is set to 0.75, which is recommended in \cite{tasel2016validated}. In general, higher values cause increased precision and decreased recall.
 \item \textbf{thick:} Sets the snake thickness. It can be set to a value between 5 and 500 (inclusive) or to ``\textbf{full}" to use all slices specified by the \verb|zrange| option. It is set to 20 by default.
 \item \textbf{cores:} Number of CPU cores utilized simultaneously for parallel processing. It is set to 1 by default.
 \item \textbf{settings-file:} Path to a YAML file to load custom settings instead of using the predefined ones. If not specified, default settings will be used.
\end{itemize}

\subsubsection{Custom Algorithm Settings}
In addition to the command line arguments and options listed above, MitoSeg requires additional settings (e.g., threshold values and iteration amounts) for fine-tuning the algorithm. MitoSeg ships with the alternative settings discussed in \cite{tasel2016validated} as three separate setting files. It allows researchers to use the supplied settings or create additional custom settings files by modifying the provided ones and then using them via the \verb|settings-file| option.

\subsubsection{Phase Selection}
As explained previously, MitoSeg runs in three phases, each focusing on a different segmentation stage. Each phase generates its own intermediate output files, which are utilized by consequent phases. If desired, only a specific phase can be executed to experiment with different settings and observe intermediate results before moving on to the next phase.

\subsubsection{Multithreaded Execution}
Most modern computers are equipped with multiple CPU cores; instead of using only a single core, MitoSeg can employ multiple threads to parallelize the segmentation operations during the first and second phases, thereby enhancing overall time performance. In the first phase, since each slice can be processed independently, the parallelization is realized on a slice-by-slice basis. In the second phase, multiple snake outputs can be extracted independently; therefore, the parallelization is performed on a snake-by-snake basis.

\subsubsection{3D Mesh Outputs for External Tools}
In the third phase, in addition to the final boundary images, MitoSeg also exports the generated outputs as .ply and .mod files, which can be displayed using 3D mesh utilities (e.g., MeshLab\footnote{\url{https://www.meshlab.net/}}) and IMOD\footnote{\url{https://bio3d.colorado.edu/imod/}}, respectively.\\

\subsubsection{MitoSeg as a Docker Application}
MitoSeg has been primarily developed and tested on GNU/Linux-based operating systems. For researchers using other operating systems, a suitable environment for MitoSeg can be set up using Docker\footnote{\url{https://www.docker.com/}}. For this purpose, a Dockerfile for building MitoSeg, along with execution scripts that manage runtime options and input/output files, are provided along with the MitoSeg source code. The provided execution scripts \verb|docker-mitoseg.sh| (for Linux/Mac) and \verb|docker-mitoseg.cmd| (for Microsoft Windows) include explanatory comments to guide users in modifying the scripts according to their specific requirements.
  
\section{Illustrative examples}
For demonstration purposes, MitoSeg was used to identify mitochondria in the \textit{gap18} dataset (accession number: 8747) \cite{perkins2003three,perkins2017microscopy} from Cell Centered Database \footnote{\url{https://library.ucsd.edu/dc/collection/bb5940732k}}. Detection and segmentation results for different datasets are presented and discussed in \cite{tasel2016validated}.\\

The tests were conducted on a desktop PC running a 64-bit Ubuntu 22.04 GNU/Linux system with a 12-cores Intel Core i7-8700 CPU and 16 GB RAM. MitoSeg was executed on slices 35 through 74 of the aforementioned dataset with the \verb|psize| option set to 2.2nm/px, while the rest of the options and settings were kept at their default values. Each image in the dataset is a 350x600 PNG image, with the total size of the selected slices being 7.0 MB.\\

The tests were repeated multiple times to observe the elapsed time and memory usage for each phase individually (with the use of \verb|phase| option) and for all phases altogether. The results of these tests are presented in Table \ref{table:table2}.

\begin{table}[ht]\centering
\caption{Memory usage and total duration for test execution}\label{table:table2}
\begin{tabular}{r|rr}
                      & \textbf{Elapsed Time} & \textbf{Peak Memory Usage} \\\hline
\textbf{Phase 1 Only} & 10.61 secs            & 246.2 MB                   \\
\textbf{Phase 2 Only} & 66.55 secs            & 907.3 MB                   \\
\textbf{Phase 3 Only} & 4.06 secs             & 683.6 MB                   \\
\textbf{All Phases}   & 76.35 secs            & 907.3 MB                  
\end{tabular}
\end{table}

\begin{figure}[ht]
\centering
{\phantomsubcaption\label{fig:fig2a}\phantomsubcaption\label{fig:fig2b}\phantomsubcaption\label{fig:fig2c}\phantomsubcaption\label{fig:fig2d}}
\includegraphics[width = .9\linewidth]{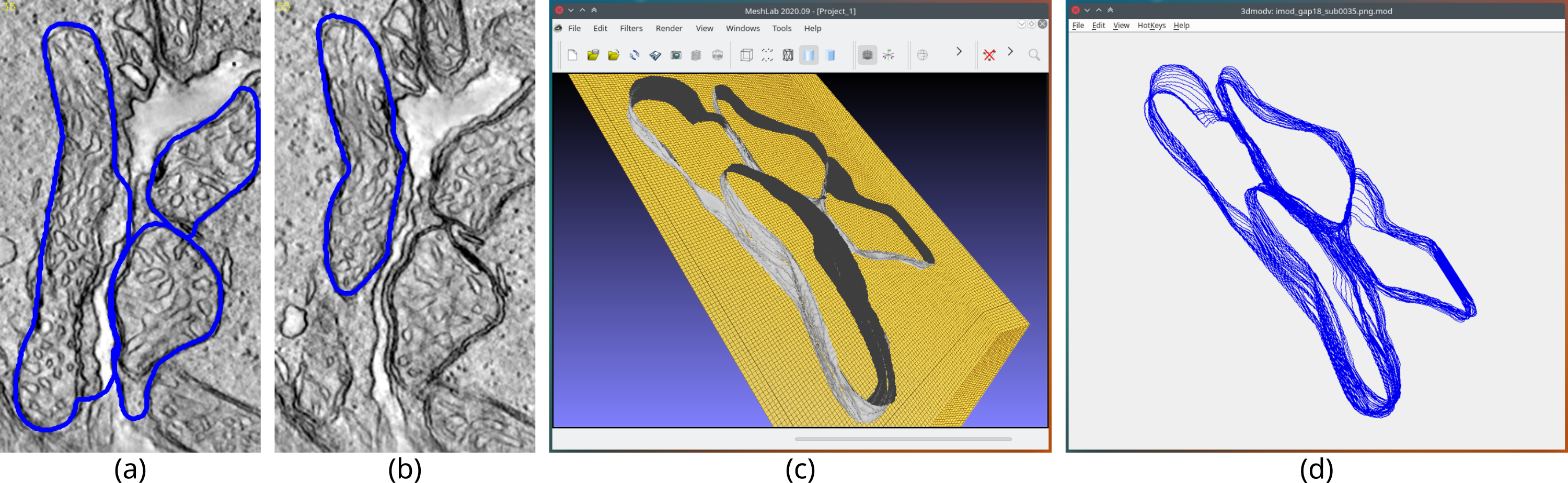}
\caption{Outputs. (a) Boundaries from slice 35, (b) boundaries from slice 55, (c) final output (.ply file), (d) final output (.mod file).}\label{fig:fig2}
\end{figure}

Since the thickness value was kept at its default of 20, the final output contains two layers of extracted mitochondria boundaries: the first is from slices 35 to 54, and the second is from slices 55 to 74. Figure \ref{fig:fig2} shows sample outputs, with the first two subfigures displaying the extracted boundaries from the dataset slices and the last two presenting the produced .ply and .mod outputs as viewed in MeshLab and IMOD, respectively.

\begin{figure}[ht]
\centering
{\phantomsubcaption\label{fig:fig3a}\phantomsubcaption\label{fig:fig3b}}
\includegraphics[width = .85\linewidth]{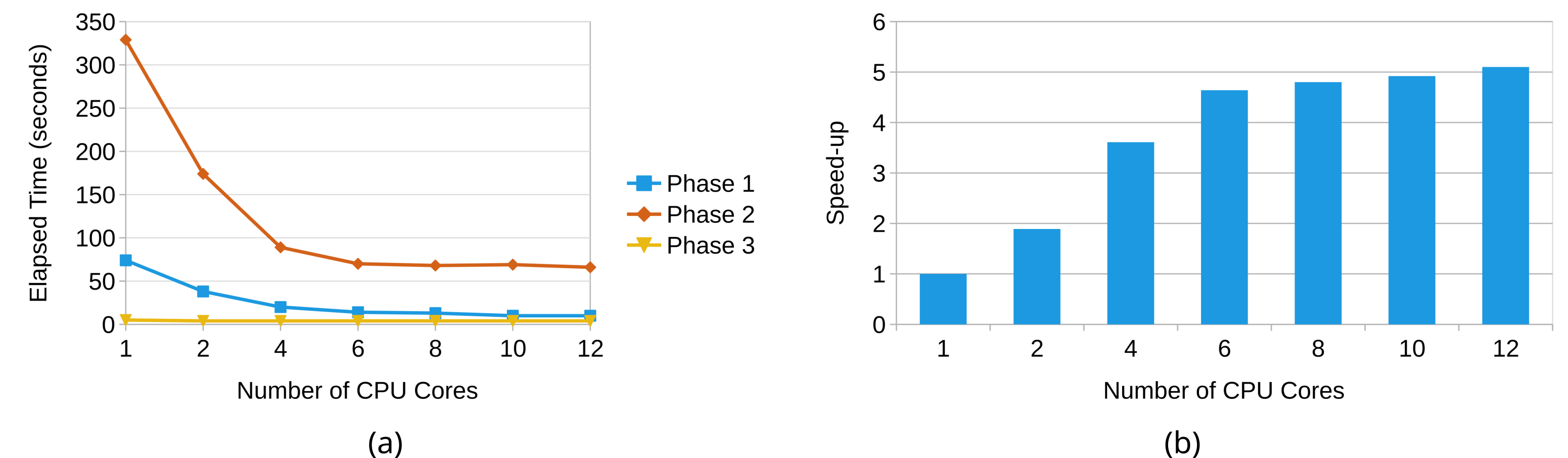}
\caption{Number of utilized CPU cores vs (a) time performance, (b) speed-up.}\label{fig:fig3}
\end{figure}

The effects of core utilization on execution time have also been tested. For this demonstration, the total duration of each phase has been recorded for the increasing number of CPU cores utilized. The time performance of MitoSeg for the \textit{gap18} dataset (using the same parameters as the previous experiment) is presented in Figure \ref{fig:fig3a}. It is observed that there is a significant decrease in the execution times for up to 6 cores in the first and second phases in which parallelization operations are performed. Note that these results highly depend on the number of mitochondria and the size of the dataset provided as input. It achieved a 5.1x speed-up in terms of overall execution time by utilizing all 12 cores compared to single-core execution, as illustrated in Figure \ref{fig:fig3b}.

\section{Impact}
In the modern medical field, there is growing interest in understanding the connection between neurodegenerative diseases and mitochondrial structure. Several mitochondrial characteristics, including the thickness of the inner and outer membranes, the structure and number of cristae, the crista junctions, and the size of contact sites, have been measured and suggested to influence mitochondrial function \cite{perkins2003three}. In this context, MitoSeg plays a crucial role in distinguishing mitochondrial regions from non-mitochondrial ones, serving as a foundation for methods aimed at studying the internal structure of mitochondria since it is able to process non-condensed mitochondrial images. Additionally, the curve-fitting technique employed by MitoSeg can be considered to assist in analyzing crista structures \cite{tasel2016validated}.\\

MitoSeg is developed as a fully automated, user-friendly tool to address the challenge above. Its features, listed in the software functionalities section, make it highly flexible and increase MitoSeg's usability. The source code is openly available for researchers, including biologists and computer scientists, who wish to use or modify the methods implemented in MitoSeg.\\

MitoSeg is particularly important in generating datasets containing isolated mitochondrial regions. Mitochondrial images often contain other cellular structures, like the endoplasmic reticulum, in addition to mitochondria. By eliminating these structures, MitoSeg aids in creating new datasets that can be used in approaches requiring training, such as CNNs, for the segmentation of mitochondrial internal structures and exploring their links to diseases.\\

Since MitoSeg uses features from the general physical structure of mitochondria in its segmentation method, it can be quickly adapted to work with images obtained with new modalities as well as TEM / SBFSEM imaging techniques without the need for a training phase (and therefore a training dataset). It is also possible to adapt MitoSeg to be used with mitochondria images obtained via different preparation techniques (e.g., condensed mitochondria images).\\

The software integrates the careful implementation of sophisticated methods. The algorithm that is used by MitoSeg and presented in \cite{tasel2016validated} has inspired various studies to date, including membrane segmentation \cite{siggel2024colabseg} and segmentation using CNNs \cite{xiao2018automatic,xiao2018effective}. Furthermore, MitoSeg has the potential to support future research in the identification and segmentation of other intracellular structures.

\section{Conclusions}

In this paper, we have presented MitoSeg, a utility designed for detecting and segmenting mitochondria boundaries in EMT images. MitoSeg integrates preprocessing, ridge detection, energy mapping, curve fitting, shape extraction, validation, and postprocessing into three distinct phases. Through these steps, our evaluations demonstrate that MitoSeg is capable of accurately detecting mitochondria, even when applied to unprocessed raw datasets that suffer from low contrast, oversampling, or noise.\\

MitoSeg also provides users with several options that enable the segmentation process to be fine-tuned. In addition to generating 2D output images, it can produce 3D mesh representations. The utility supports multicore systems, thereby reducing execution time, and is accessible across major operating systems through a Docker environment.\\

Tomographic images produced by TEM are typically characterized by relatively low thickness along the z-axis compared to the x and y axes, making pseudo-3D segmentation approaches suitable for TEM images. However, such approaches can introduce discontinuities between independently processed layers, as observed in Figures \ref{fig:fig2c} and \ref{fig:fig2d}. To address this issue, the segmentation process can be refined by enhancing the balloon snake model to connect segmented mitochondria contours on adjacent layers, followed by re-executing the fitting algorithm. For other imaging modalities, such as SBFSEM and cryo-EM, fully 3D segmentation models can be directly employed.\\

Future enhancements of MitoSeg may include adapting the model to GPU parallelization, given the suitability of the balloon snake model for such optimizations.

\section*{Acknowledgments}\label{}
The authors would like to express their gratitude to Erkan U. Mumcuoglu, Reza Z. Hassanpour, and Guy Perkins for their support in algorithm conceptualization, clarification of biological background and significance, and for providing the datasets.

\newpage



\bibliographystyle{elsarticle-num} 
\bibliography{refs}




%
%
%

\end{document}